\documentclass[conference]{IEEEtran}
\IEEEoverridecommandlockouts
\usepackage[normalem]{ulem}
\usepackage{cite}
\usepackage{amsmath,amssymb,amsfonts}
\usepackage{algorithmic}
\usepackage{graphicx}
\usepackage{textcomp}
\usepackage{xcolor}
\usepackage{multirow}
\usepackage{algorithm}
\def\BibTeX{{\rm B\kern-.05em{\sc i\kern-.025em b}\kern-.08em
    T\kern-.1667em\lower.7ex\hbox{E}\kern-.125emX}}
\begin{document}

\title{Online Test-Time Adaptation for Generalizable Dynamic Graph Anomaly Detection}

\author{
\IEEEauthorblockN{1\textsuperscript{st} Jialun Zheng}
\IEEEauthorblockA{
\textit{Advanced Interdisciplinary Science}\\
\textit{and Technology}\\
\textit{Zhejiang University of Technology}\\
Hangzhou, Zhejiang, China\\
zhengjialun1999@gmail.com
}
\and
\IEEEauthorblockN{2\textsuperscript{nd} Hanchen Yang}
\IEEEauthorblockA{
\textit{Department of Computing}\\
\textit{The Hong Kong Polytechnic University}\\
Hong Kong, China\\
hanchen.yang@connect.polyu.hk
}
\and
\IEEEauthorblockN{3\textsuperscript{rd} Jiannong Cao}
\IEEEauthorblockA{
\textit{Department of Computing}\\
\textit{The Hong Kong Polytechnic University}\\
Hong Kong, China\\
jiannong.cao@polyu.edu.hk
}
\and
\IEEEauthorblockN{4\textsuperscript{th} Yankai Chen\textsuperscript{*}}
\IEEEauthorblockA{
\textit{Department of Computer Science}\\
\textit{University of Illinois at Chicago (UIC)}\\
Chicago, IL, USA\\
yankaichen@acm.org
}
\and
\IEEEauthorblockN{5\textsuperscript{th} Yuanjing Feng\textsuperscript{*}}
\IEEEauthorblockA{
\textit{Institute of Information Processing}\\
\textit{and Automation}\\
\textit{Zhejiang University of Technology}\\
Hangzhou, Zhejiang, China\\
fyjing@zjut.edu.cn
}
\and
\IEEEauthorblockN{6\textsuperscript{th} PHILIP S. YU}
\IEEEauthorblockA{
\textit{Department of Computer Science}\\
\textit{University of Illinois at Chicago (UIC)}\\
Chicago, IL, USA\\
psyu@uic.edu
}
\thanks{*Corresponding authors: Yankai Chen
(yankaichen@acm.org) and Yuanjing Feng
(fyjing@zjut.edu.cn).}
}

\maketitle

\begin{abstract}
Generalizable dynamic graph anomaly detection (DGAD) enables pretrained detectors to identify anomalies in unseen target domains
without costly retraining. However, existing methods assume access to the full target-domain data, whereas in practice target data
arrive sequentially in unlabeled chunks, which is the online test-time adaptation (OTTA) setting. Under this setting, each chunk
provides only a partial view of the graph, making it difficult to preserve historical context. To address the aforementioned limitations, we formulate online
test-time adaptation for generalizable DGAD and propose OTTA-DGAD. OTTA-DGAD first extracts dynamic prototypes, i.e., evolving
representations of normal and anomalous patterns, from temporal ego-graphs and stores them in a memory buffer. The buffer
selectively retains general patterns shared across the source domains used for pretraining while incorporating new patterns from the
target domain. An anomaly scorer then compares incoming edge representations against these prototypes to identify both general
and domain-specific anomalies. During adaptation, it updates the memory buffer only with reliable pseudo-labels, ranked by a score
that combines prediction entropy with structural support from neighboring edges and is normalized within each predicted class.
It further introduces cross-chunk context enrichment, which retrieves representative representations retained from earlier chunks
to compensate for information lost as data arrive sequentially. Extensive experiments under strict test-then-adapt OTTA settings
demonstrate state-of-the-art performance on ten real-world datasets from diverse domains.
\end{abstract}


\section{Introduction}
Graph anomaly detection (GAD) aims to identify unusual nodes or edges from their features and structural relations. It has become increasingly important in applications such as fraud detection, transaction monitoring, and social-network security~\cite{deng2022graph,wang2025epm,pan2025label,zhang2025sparsefraudnet}. In practice, however, many graphs are evolving rather than static~\cite{zheng2024inductive,dygmf,he2023dynamically,zheng2025coin}. For example, new users and items continuously join a platform, and user--content interactions evolve over time. This motivates dynamic graph anomaly detection (DGAD), which captures not only abnormal patterns, but also how such abnormality evolves over time~\cite{liu2021anomaly,chen2024fine,lee2024slade}. In online deployment, target data arrive continuously in unlabeled chunks rather than becoming available all at once~\cite{gong2023sotta,yuan2024tea}. A detector pretrained on source domains must first predict anomalies in each incoming chunk and then adapt using the current chunk, without accessing target labels or future chunks. This is the online test-time adaptation (OTTA) setting~\cite{liang2025comprehensive,wang2025search,karmanov2024efficient,ai2026promptdyg} considered in this work.

Existing DGAD methods~\cite{wang2025epm,liu2021anomaly} often use temporal representations to detect anomalies by measuring deviations from normal patterns. However, this one-model-per-domain paradigm~\cite{yang2025generalizable,chen2024fine,lee2024slade,su2024cross} generalizes poorly to new domains with different anomaly patterns, feature distributions, and graph structures, namely domain shift~\cite{qiaogcal,tian2024freedyg,zhang2024disentangling}. A straightforward solution is to retrain the detector for each new domain. Nevertheless, this is costly and difficult to scale, especially when new domains continuously emerge~\cite{zheng2025coin,zheng2024inductive}. Recent generalist GAD methods~\cite{liu2024arc,qiao2025anomalygfm,niu2024zero} improve cross-domain generalization through pretraining on multiple source datasets. Related approaches include GUDI~\cite{li2024graph}, which combines domain-agnostic pretraining with few-shot adaptation for graph anomaly detection, and CPDG~\cite{bei2024cpdg}, which uses structural-temporal contrastive pretraining to improve the transferability of dynamic graph representations. These approaches explore transferable graph representations, while adaptation to sequentially arriving unlabeled target chunks requires further investigation. More directly related to generalizable DGAD, DP-DGAD~\cite{zheng2026dp} extracts dynamic prototypes from temporal ego-graphs to capture evolving normal and anomalous patterns across domains. Its prototype memory retains domain-agnostic patterns while incorporating domain-specific ones, and high-confidence target predictions provide pseudo-labels for adaptation. These mechanisms provide a foundation for generalizable DGAD. However, extending such adaptation to the OTTA setting poses two key challenges.

\begin{figure}[t]
\centerline{\includegraphics[width=0.95\linewidth]{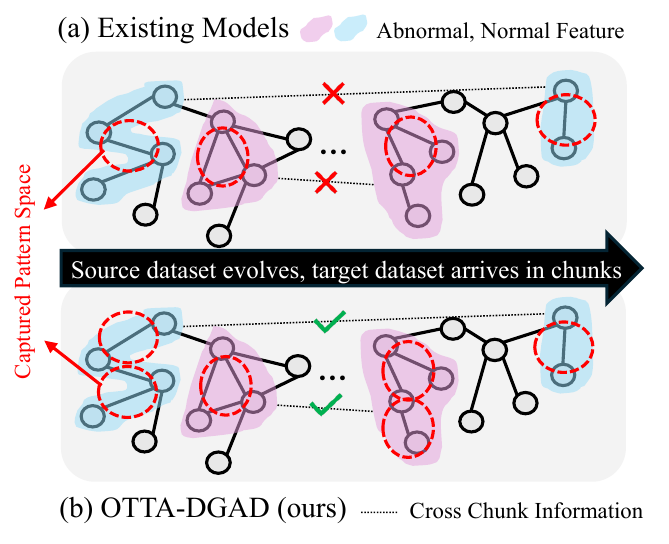}}
  \caption{Motivating illustration of cross-chunk context enrichment. (a) Direct chunk-wise processing may miss historical context relevant to current anomaly detection. (b) Our method, OTTA-DGAD, preserves and retrieves historical ego-graph representations to enrich current-chunk representations.}
  \label{teaser}
\end{figure}

The first major challenge is incomplete historical context under chunk-wise observations. In dynamic graphs, each incoming chunk only provides a partial view of the target domain. However, many anomaly patterns depend on cross-chunk interactions, historical neighbors, and evolving structural dependencies. Such information is not directly observable from the current chunk~\cite{sun2024program,cai2026continual}. Take a web platform as an example, where interaction logs arrive in hourly or daily batches. A user may interact with only a few items or pages within one chunk, and collusive or spam behavior may become evident only after linking multiple chunks. Consequently, directly processing each chunk may miss relevant historical interactions and lead to incorrect anomaly detection, as illustrated in Fig.~\ref{teaser}(a). Although DP-DGAD retains normal and anomalous patterns in its prototype memory, such class-level patterns do not directly provide the historical context missing from the current edge's temporal ego-graph. Existing OTTA methods can also track historical scores or gradients~\cite{marsden2024universal,duan2026lifelong}, but do not explicitly retrieve historical graph representations to enrich current-chunk representations. Therefore, the model needs to preserve and retrieve cross-chunk context, as shown in Fig.~\ref{teaser}(b).

Secondly, the unlabeled nature of the online target stream makes reliable prototype updating challenging~\cite{fuchs2025online,niu2023towards,bar2024protected}. Anomaly patterns evolve over time both between and within domains~\cite{wan2024federated,qiaogcal,tian2024freedyg,zhang2024disentangling}. A pattern that appears abnormal at an early stage may later become common as the target domain evolves. Together with incomplete historical context, these changes may cause the detector to produce incorrect predictions even with high confidence. DP-DGAD selects confident target predictions based on low prediction entropy for adaptation. However, prediction confidence alone does not indicate whether a pseudo-label is supported by the surrounding graph structure. Incorrect pseudo-labels may therefore enter the prototype memory and affect anomaly detection in subsequent chunks. Moreover, reliability scores for pseudo-labeled normal and anomalous edges may have different scales, introducing bias when selecting samples for adaptation. Therefore, the model needs to combine prediction confidence with structural support and account for class-dependent score differences when updating prototypes.

To address these challenges, we propose Online Test-Time Adapted Generalizable Dynamic Graph Anomaly Detector (OTTA-DGAD). Building on dynamic prototype-based detection, OTTA-DGAD is pretrained on multiple source domains and uses prototypes to represent evolving normal and anomalous patterns. A prototype memory retains transferable knowledge while incorporating target-specific patterns through online updates. As for missing historical context, we introduce cross-chunk context enrichment, which stores representative and diverse ego-graph representations and retrieves relevant context to enrich representations in subsequent chunks. Unlike the prototype memory used for anomaly scoring, this context buffer supplements current representations with historical information. For reliable prototype updates, we combine prediction entropy with structural support from neighboring edges. The resulting reliability scores are normalized within each pseudo-label class and globally ranked to select reliable target edges for memory adaptation. For each target chunk, anomaly prediction is performed before updating the prototype memory, following the strict test-then-adapt protocol. Our main contributions are summarized as follows:
\begin{itemize}
    \item[$\bullet$] We formulate generalizable dynamic graph anomaly detection under online test-time adaptation settings. The pretrained DGAD model must predict anomalies before adapting to each sequentially arriving unlabeled target chunk, without accessing target labels or future chunks.
    \item[$\bullet$] We introduce cross-chunk context enrichment, a module that stores representative and mutually distinct ego-graph representations from earlier chunks and retrieves relevant historical context to enrich current-chunk representations.
    \item[$\bullet$] We combine structural support and prediction entropy for class-wise normalized reliability ranking, allowing reliable target pseudo-labels to guide prototype updates.
    \item[$\bullet$] We conduct extensive experiments on ten real-world datasets under a strict OTTA protocol. The results show that our method achieves state-of-the-art performance with improved stability and generalization.
\end{itemize}
\section{RELATED WORK}

\subsection{Generalizable Dynamic Graph Anomaly Detection}\label{AA}

Existing dynamic graph anomaly detection (DGAD) methods jointly model structural and temporal dependencies. TADDY \cite{liu2021anomaly} employs a transformer to encode spatial dependencies and temporal evolution, while StrGNN and DynAnom \cite{cai2021structural,guo2022subset} investigate local structural patterns in evolving graphs. Other representative methods include semi-supervised SAD \cite{tian2023sad}, fine-grained temporal modeling in FALCON \cite{chen2024fine}, and self-supervised learning from edge streams in SLADE \cite{lee2024slade}. For multiplex graphs, UMGAD~\cite{li2025umgad} combines node attribute and graph structure reconstruction with contrastive learning to detect anomalous nodes across multiple interaction types. GeneralDyG \cite{yang2025generalizable} further explores generalizable anomaly detection in dynamic graphs. However, adapting pretrained detectors to sequentially arriving, unlabeled target chunks remains challenging, particularly when anomaly patterns and structural context evolve over time.

Recent generalist graph anomaly detectors, including ARC \cite{liu2024arc}, AnomalyGFM \cite{qiao2025anomalygfm}, and UNPrompt \cite{niu2024zero}, aim to transfer detection knowledge across graphs. Related strategies include cross-domain graph pretraining \cite{zhao2024all}, normalization-based anomaly detection \cite{li2023zero}, and graph prompting \cite{liu2023graphprompt}. These studies provide foundations for cross-domain transfer, but do not directly resolve online adaptation to chunked dynamic graph streams. In this setting, detectors must retain transferable knowledge while incorporating evolving target-specific patterns. OTTA-DGAD addresses this challenge through dynamic prototype memory and cross-chunk context enrichment.

\subsection{Online Test-Time Adaptation for Dynamic Graph Anomaly Detection}

Online test-time adaptation (OTTA) adapts a source-pretrained model to sequentially arriving, unlabeled target data without accessing source data during adaptation \cite{liang2025comprehensive,wang2025search}. Existing methods improve adaptation stability through controlled updates, sample selection, and memory mechanisms. For example, CMF \cite{lee2024continual} applies momentum filtering in parameter space, SoTTA \cite{gong2023sotta} selects reliable samples from noisy streams, and Protected TTA \cite{bar2024protected} uses online entropy matching to guard against harmful updates. Other approaches explore energy-based adaptation \cite{yuan2024tea}, ensemble-based correction \cite{marsden2024universal}, complementary memories \cite{zhang2024dual}, and lightweight adaptation for vision-language models \cite{karmanov2024efficient}. Although these strategies improve robustness under distribution shifts, they are not specifically designed to recover historical structural context missing from individual graph chunks.

Related studies also investigate memory mechanisms and efficient representation computation on evolving graphs. DyFMVP~\cite{pang2025dyfmvp} addresses memory staleness in continuous-time dynamic graphs, while incremental GNN embedding computation~\cite{wang2026incremental} reuses historical computation results to reduce redundant computation on streaming graphs. These studies provide insights into maintaining and reusing historical information, while our work focuses on retrieving historical ego-graph representations and reliably updating prototypes under unlabeled online target streams. For dynamic graphs, PromptDyG \cite{ai2026promptdyg} adapts lightweight prompts while retaining a frozen pretrained backbone. Related online adaptation approaches include ADCSD \cite{guo2025online}, which combines short-term correction with a long-term historical state for spatial--temporal forecasting, and LCoTTA \cite{duan2026lifelong}, which constrains updates using an online-tracked gradient subspace. These mechanisms capture different forms of adaptation history, but do not explicitly retrieve historical graph representations to enrich current-chunk anomaly detection. In contrast, OTTA-DGAD combines dynamic prototype memory, entropy--structural reliability ranking, and cross-chunk context retrieval to preserve transferable patterns while incorporating reliable target-specific information.
\section{PRELIMINARIES}

\textbf{Definition 1. (Dynamic Graph Anomaly Detection)} A dynamic graph can be denoted as $\mathcal{G} = \{G_1, G_2, \ldots, G_T\}$, where $T$ is
the number of intervals, and interval $t$ consists of $M_t$ timestamps. Specifically,
$G_t = (V_t, E_t)$, where $V_t$ is a node set and $E_t$ is an edge set, with
$A^{V_t} \in \mathbb{R}^{N^{V_t} \times N^{V_t}}$ and
$A^{E_t} \in \mathbb{R}^{N^{E_t} \times N^{E_t}}$ denoting the node and edge adjacency
matrices ($N^{V_t}=|V_t|$, $N^{E_t}=|E_t|$), and
$X^{V_t} \in \mathbb{R}^{N^{V_t} \times D^V \times M_t}$,
$X^{E_t} \in \mathbb{R}^{N^{E_t} \times D^E \times M_t}$ denoting the temporal node and
edge features, where $D^E$ and $D^V$ are the edge and node feature dimensions, respectively. The edges at interval $t$ are associated with a label vector $y_t \in \{0,1\}^{N^{E_t}}$,
partitioning $E_t$ into the normal edge set $E_t^n$ and the abnormal edge set $E_t^a$.

Following existing online test-time adaptation methods~\cite{wang2025search}, we further divide $\mathcal{G}$ into $Q$ data chunks $\{\mathcal{C}_1, \mathcal{C}_2,
\ldots, \mathcal{C}_Q\}$ for online processing. Each chunk $\mathcal{C}_q = (V_q, E_q)$ aggregates $M_q$ consecutive timestamps and contains a fixed number of
edges $|E_q| = B$ (e.g., $B=128$ interactions). $\mathcal{C}_q$ inherits the
same notation as $G_t$ (i.e., $A^{V_q}$, $A^{E_q}$, $X^{V_q}$, $X^{E_q}$, $E_q^n$, $E_q^a$,
$y_q$). The goal of dynamic graph anomaly detection is to distinguish abnormal edges from
normal ones based on the evolving graph structure and temporal features.

 \textbf{Definition 2. (Continuous Domain Shift)} Consider a collection $\{\mathcal{G}_1, \mathcal{G}_2, \ldots, \mathcal{G}_{S+K}\}$ of $(S+K)$ dynamic graphs from different domains, where $\{\mathcal{G}_{1}, \mathcal{G}_{2}, \ldots, \mathcal{G}_{S}\}$ are from source domains and $\{\mathcal{G}_{S+1}, \mathcal{G}_{S+2}, \ldots, \mathcal{G}_{S+K}\}$ are from target domains. For two dynamic graphs from different domains, $\mathcal{G}_{i}$ and $\mathcal{G}_{j}$ where $i \neq j$, we define domain shift as a difference in their node and edge feature distributions, graph structures (such as adjacency matrices), or mappings from edges to anomaly labels. Such domain shifts occur among any two dynamic graphs in the dataset collection $\{\mathcal{G}_1, \mathcal{G}_2, \ldots, \mathcal{G}_{S+K}\}$.

 \textbf{Problem Definition. (Online Test-Time Adaptation for Generalizable Dynamic Graph Anomaly Detection)}
Given $S$ source dynamic graphs $\{\mathcal{G}_1, \ldots, \mathcal{G}_S\}$, we first train
a source detector $\Psi_0$ that distinguishes abnormal from normal edges on source dynamic graphs. Given an unseen
target dynamic graph $\mathcal{G}_k$ partitioned into $Q_k$ chunks $\{\mathcal{C}_1,
\ldots, \mathcal{C}_{Q_k}\}$, our goal is to sequentially adapt $\Psi_0$ into a series of
detectors $\{\Psi_1, \ldots, \Psi_{Q_k}\}$ that minimize the anomaly detection error across
all chunks of $\mathcal{G}_k$, without access to any target label. Specifically, for each chunk $\mathcal{C}_q$, the detector $\Psi_{q-1}$ adapted using the previous chunk $\mathcal{C}_{q-1}$ first predicts anomaly scores on the
current unlabeled chunk $\mathcal{C}_q$ as:

\begin{equation}
\hat{y}_{q} = \Psi_{q-1}\!\left(A^{V_{q}}, A^{E_{q}}, X^{V_{q}}, X^{E_{q}}\right),
\end{equation}

and is then updated to $\Psi_q$ using only $\mathcal{C}_q$ and $\hat{y}_q$, without
access to $y_q$ or future chunks.
\section{METHODOLOGY}

\begin{figure*}[t]
	\centering
	\includegraphics[width=\textwidth]{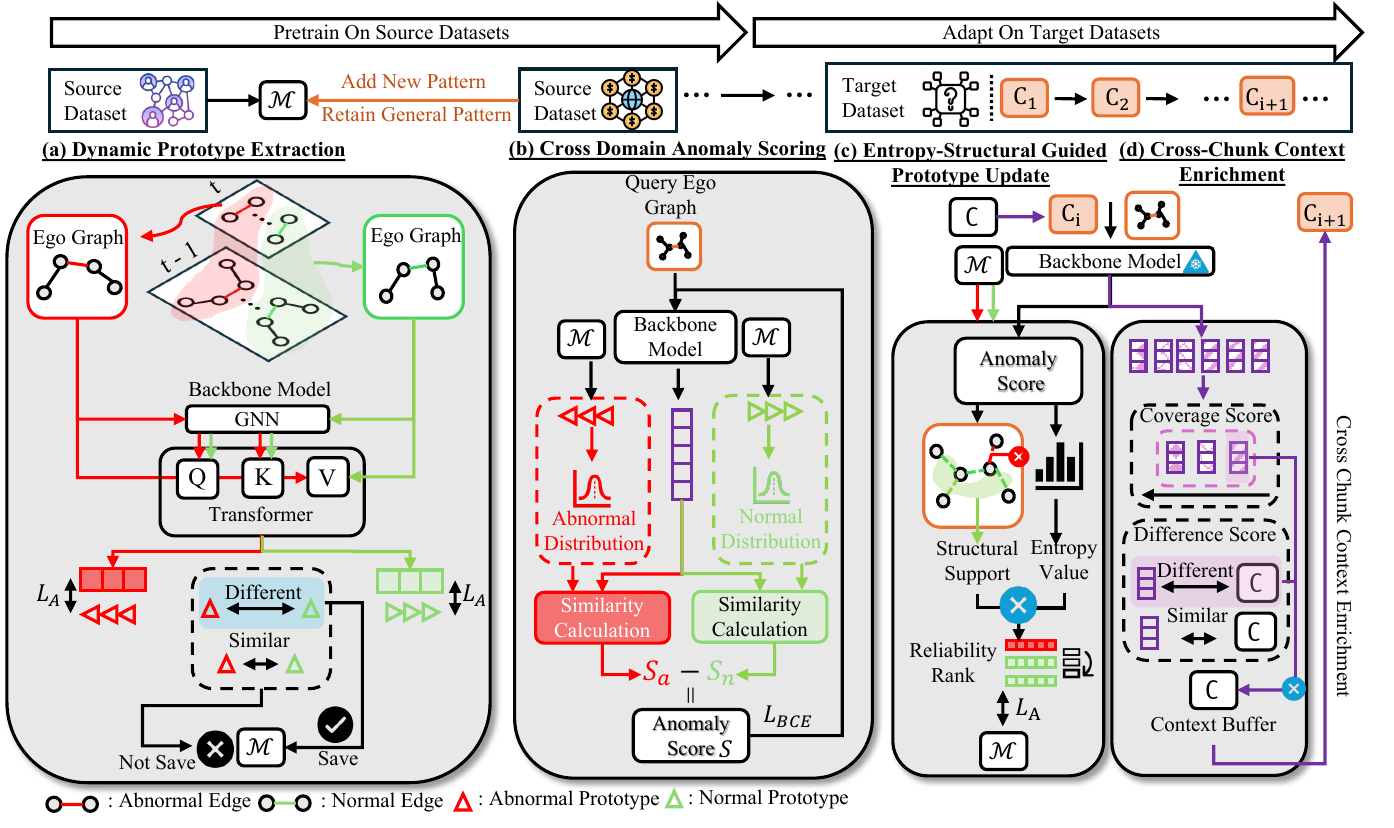}
	\caption{General framework of OTTA-DGAD. (a) Starting with the first source dataset, we extract ego-graphs to capture temporal patterns and store the most distinct prototype pairs in a memory buffer. (b) As we pretrain on the following source datasets, new domain-specific patterns are added while general domain-agnostic patterns are retained. Prototype distributions are then compared with edge embeddings for anomaly scoring. (c) On target chunks, reliable pseudo-labeled samples, selected based on prediction entropy and agreement with neighboring edges, are used for memory updates. (d) The buffer retains representations with high similarity to the chunk distribution, as measured by the coverage score, and high dissimilarity to stored entries. These representations enrich subsequent chunks with cross-chunk context, while the generalist model remains frozen.}
	\label{frame}
\end{figure*}

As shown in Fig.~\ref{frame}, OTTA-DGAD first extracts evolving patterns using dynamic prototype extraction and stores them in a memory buffer. Then, during cross-domain anomaly scoring, it retains general patterns shared among pretrained source datasets while incorporating domain-specific ones. These dynamic prototypes are compared with embeddings to calculate anomaly scores. On target data, it evaluates detections using a reliability score that combines prediction entropy and structural support, where structural support reflects agreement between an edge and its neighbors. These detections provide pseudo-labels for updating the memory buffer. In addition, a context buffer stores representations to provide cross-chunk information. We select samples using two scores. The coverage score measures similarity to the distribution of the current chunk, while the difference score measures dissimilarity from entries already stored in the buffer. When a new chunk arrives, its query representations are enriched with cross-chunk information from this buffer.
\subsection{Dynamic Prototype Extraction}

Following the dynamic prototype-based formulation of
DP-DGAD~\cite{zheng2026dp}, we represent normal and anomalous
temporal patterns with paired prototypes. For an edge $e_i$
occurring at timestamp $\tau$, let $G_{\mathrm{ego},i}$ denote its
$h$-hop temporal ego-graph, where $h$ is the neighborhood hop radius
and only interactions observed no later than $\tau$ are included.
A GNN encodes local edge relations, ego-graph mean-centering captures
edge-specific deviations, and a transformer aggregates the resulting
representations. Including the projection to prototype space, the
encoder produces
\begin{equation}
z_i=f_\theta(G_{\mathrm{ego},i})\in\mathbb{R}^{d_p},
\label{eq:encoder}
\end{equation}
where $\theta$ denotes the encoder parameters and $d_p$ is the shared
dimension of edge representations and prototypes.

Let $c\in\{n,a\}$ index the normal and anomalous classes,
respectively. The prototype $\mathbf{v}_c\in\mathbb{R}^{d_p}$ is
aligned with source edges in class $c$ through
\begin{equation}
L_A=\sum_{c\in\{n,a\}}\sum_{e_i\in E_t^c}
\|z_i-\mathbf{v}_c\|_2^2,
\label{eq:alignment}
\end{equation}
where $E_t^n$ and $E_t^a$ are the normal and anomalous edge sets
at source interval $t$, and $\|\cdot\|_2$ denotes the Euclidean norm.

The prototype memory $\mathcal{B}$ has capacity $\mathcal{M}$,
measured in normal--anomalous prototype pairs. During pretraining
on the first source graph, each pair is ranked by its separation:
\begin{equation}
s_d=\frac{1}{d_p}\sum_{r=1}^{d_p}
\left|v_a^{(r)}-v_n^{(r)}\right|,
\label{eq:separation}
\end{equation}
where $v_c^{(r)}$ is coordinate $r$ of $\mathbf{v}_c$.
High-scoring pairs are retained, replacing lower-scoring pairs when
the memory is full. The highest-ranked retained pair initializes
the prototypes for the next epoch. Subsequent source graphs use
the cross-domain retention criterion below.

\subsection{Cross-Domain Anomaly Scoring}

The memory retains transferable source patterns while incorporating
patterns from subsequent source graphs. For each stored pair, its
class-matched distance $s_e$ and retention score $s_r$ are
\begin{equation}
\begin{aligned}
s_e&=\frac{1}{|E_t^n|}\sum_{e_i\in E_t^n}\|z_i-\mathbf{v}_n\|_2
+\frac{1}{|E_t^a|}\sum_{e_i\in E_t^a}\|z_i-\mathbf{v}_a\|_2,\\
s_r&=\lambda_d s_d-\lambda_e s_e,
\end{aligned}
\label{eq:retention}
\end{equation}
where $\lambda_d$ and $\lambda_e$ are nonnegative weights balancing
class separation and cross-domain matching. A high $s_r$ favors
pairs that separate the classes and match the new source graph,
guiding memory retention and prototype initialization.

At update iteration $k$, let $\mathcal{B}^{(k)}$ be the current
prototype memory and $m_k=|\mathcal{B}^{(k)}|\leq\mathcal{M}$ its
number of pairs. For a nonempty memory, let
$\mathbf{v}_{c,j}^{(k)}$ denote the class-$c$ prototype in pair
$j\in\{1,\ldots,m_k\}$. The scorer maintains a mean vector
$\mu_c^{(k)}\in\mathbb{R}^{d_p}$ and covariance matrix
$\Sigma_c^{(k)}\in\mathbb{R}^{d_p\times d_p}$ using
exponential moving averages:
\begin{equation}
\begin{aligned}
\mu_c^{(k)}&=\alpha\mu_c^{(k-1)}
+\frac{1-\alpha}{m_k}\sum_{j=1}^{m_k}\mathbf{v}_{c,j}^{(k)},\\
\Sigma_c^{(k)}&=\alpha\Sigma_c^{(k-1)}
+\frac{1-\alpha}{\max(1,m_k-1)}(C_c^{(k)})^T C_c^{(k)}.
\end{aligned}
\label{eq:statistics}
\end{equation}
Here, $\alpha\in[0,1)$ is the momentum and
$C_c^{(k)}\in\mathbb{R}^{m_k\times d_p}$ is the centered prototype matrix, whose row $j$ is equal to
$(\mathbf{v}_{c,j}^{(k)}-\mu_c^{(k)})^T$.

Using the current statistics, with the iteration index omitted,
the scorer computes normal and anomalous compatibility scores
$s_{n,i}$ and $s_{a,i}$, respectively:
\begin{equation}
\begin{aligned}
s_{n,i}&=z_i^T\mu_n-\lambda_n z_i^T\Sigma_n z_i,\\
s_{a,i}&=z_i^T\mu_a-\lambda_a z_i^T\Sigma_a z_i,\\
s_i&=s_{a,i}-s_{n,i},\qquad
p_i=\frac{1}{1+\exp(-s_i)}.
\end{aligned}
\label{eq:scoring}
\end{equation}
The scalars $\lambda_n$ and $\lambda_a$ are learned during source
pretraining and control the covariance terms. The logit $s_i$
measures relative compatibility with the anomalous class, and
$p_i\in(0,1)$ is the model-predicted probability that edge $e_i$
is anomalous.

Source training combines binary cross-entropy and prototype alignment:
\begin{equation}
\begin{aligned}
L&=\lambda_{BCE}L_{BCE}+\lambda_A L_A,\\
L_{BCE}&=-\frac{1}{|E_t|}\sum_{e_i\in E_t}
\big[y_i\log p_i+(1-y_i)\log(1-p_i)\big],
\end{aligned}
\label{eq:source-objective}
\end{equation}
where $E_t=E_t^n\cup E_t^a$, $y_i\in\{0,1\}$ is the source label
($1$ for anomalous and $0$ for normal), and $\lambda_{BCE}$ and
$\lambda_A$ are nonnegative loss weights.

At target time, the encoder and learned scorer parameters remain
frozen, while the prototype memory and distribution statistics are
maintained online. Scoring uses the context-enriched representation
$\tilde{\mathbf{z}}_i\in\mathbb{R}^{d_p}$ in place of $z_i$;
its construction is defined in Section~IV-D. For each incoming
chunk, predictions precede prototype adaptation and context-buffer
updates. Sections~IV-C and IV-D describe these online modules.

\subsection{Entropy-Structural Guided Prototype Update}

To adapt the source-pretrained detector to an unlabeled target stream, we update the dynamic prototype memory using only reliable pseudo-labeled target edges. In addition to prediction entropy, we use structural support as a reliability criterion and then perform class-wise memory updates to alleviate the bias caused by imbalanced confident pseudo-labels.

Given an incoming target chunk $\mathcal{C}_q$, the pretrained anomaly scorer produces an anomaly probability $p_i \in [0,1]$ for each target edge $e_i$ based on the enhanced representation $\tilde{\mathbf{z}}_i$, together with its pseudo-label
\begin{equation}
\hat{y}_i = \mathbb{I}(p_i > 0.5),
\end{equation}
where $\hat{y}_i = 1$ and $\hat{y}_i = 0$ denote abnormal and normal pseudo-labels, respectively.

We first estimate the prediction confidence using entropy:
\begin{equation}
\mathcal{H}(p_i) = -p_i \log (p_i+\epsilon) - (1-p_i)\log (1-p_i+\epsilon),
\end{equation}
where lower entropy indicates higher prediction confidence. We further convert it into a normalized confidence score $\rho_i^{\mathrm{ent}}$:
\begin{equation}
\rho_i^{\mathrm{ent}} = 1-\frac{\mathcal{H}(p_i)}{\ln 2}.
\end{equation}

However, confidence alone is insufficient under continuous domain shift. We therefore introduce a structural support score to evaluate whether the predicted class of $e_i$ is supported by its local neighborhood. Let $\mathcal{N}_i$ denote the neighboring edges of $e_i$ in its temporal ego-graph. The structural support score $\rho_i^{\mathrm{str}}$ is defined as
\begin{equation}
\rho_i^{\mathrm{str}}=
\begin{cases}
\sum\limits_{j\in\mathcal{N}_i}(1-p_j), & \hat{y}_i=0,\\[6pt]
\sum\limits_{j\in\mathcal{N}_i}p_j, & \hat{y}_i=1,
\end{cases}
\end{equation}
\indent In this way, a pseudo-labeled normal edge is supported by normal-like neighbors, while a pseudo-labeled abnormal edge is supported by abnormal-like neighbors. Based on the above two criteria, we assign each target edge a raw reliability score:
\begin{equation}
r_i = \rho_i^{\mathrm{ent}} \cdot \rho_i^{\mathrm{str}}.
\end{equation}

Since the structural support score is computed in a class-conditional manner, the raw reliability scores of pseudo-labeled normal and abnormal edges may be on different scales. To make the scores comparable for global ranking, we further normalize them within each pseudo-label class. Specifically, for each pseudo-label class present in the current chunk, let
\begin{equation}
r_{\min}^{(y)} = \min_{\{i \mid \hat{y}_i = y\}} r_i,
\qquad
r_{\max}^{(y)} = \max_{\{i \mid \hat{y}_i = y\}} r_i,
\end{equation}
where $y \in \{0,1\}$. The class-normalized reliability score is defined as
\begin{equation}
\tilde{r}_i =
\frac{r_i - r_{\min}^{(\hat{y}_i)}}
{r_{\max}^{(\hat{y}_i)} - r_{\min}^{(\hat{y}_i)} + \epsilon}.
\end{equation}

We then select the top $N_{con}$ target edges with the highest normalized reliability scores as confident detections. According to their pseudo-labels, the selected confident detections are further divided into abnormal and normal subsets ${E}_{q}^{a}, {E}_{q}^{n}$. For each class $c\in\{n,a\}$, we construct a class-wise confident representation only when $\sum_{e_i\in E_q^c}\tilde r_i>0$, using normalized reliability-weighted aggregation:
\begin{equation}
\mathbf{z}_{q}^{a} =
\frac{\sum\limits_{e_i \in E_{q}^{a}} \tilde{r}_i \tilde{\mathbf{z}}_i}
{\sum\limits_{e_i \in E_{q}^{a}} \tilde{r}_i + \epsilon},
\qquad
\mathbf{z}_{q}^{n} =
\frac{\sum\limits_{e_i \in E_{q}^{n}} \tilde{r}_i \tilde{\mathbf{z}}_i}
{\sum\limits_{e_i \in E_{q}^{n}} \tilde{r}_i + \epsilon}.
\end{equation}

We then align each valid class-wise confident representation with the corresponding memory
prototypes through the alignment loss $\mathcal{L}_A$, leaving the prototypes of all other classes unchanged.
Singleton pseudo-label classes and classes with identical raw reliability scores receive zero normalized weights and therefore do not trigger prototype updates.
If neither selected class subset has a positive total weight, the prototype memory $\mathcal B$ and all four distribution statistics $\mu_n,\mu_a,\Sigma_n,\Sigma_a$ remain unchanged.
Thus, only reliable pseudo-labeled target edges contribute to memory adaptation. By normalizing the reliability scores within each pseudo-label class before global ranking, the proposed strategy reduces the scale bias across classes while still allowing the numbers of selected abnormal and normal detections to adapt to each target chunk.

\subsection{Cross-Chunk Context Enrichment}

The temporal ego-graph of a target edge contains temporal
and structural context, but may be incomplete within a single
chunk. As a result, the pretrained detector may generate
incorrect predictions and pseudo-labels because of insufficient
context. To address this issue, we introduce a Cross-Chunk
Context Enrichment module, which supplements current edge
representations with historical context before anomaly scoring,
reliable pseudo-label selection, and prototype adaptation.

We maintain a context buffer $\mathcal{B}_c$ with a fixed
maximum capacity $\mathcal{M}_{c,\max}$. At the beginning
of each target domain, $\mathcal{B}_c$ is initialized as empty.
It stores historical ego-graph representations rather than
raw edges or complete target chunks.  Thus, historical context can be
reused without replaying or reprocessing full historical chunks.
The context buffer is different from the prototype memory
$\mathcal{B}$. $\mathcal{B}$ stores normal and anomalous
prototypes for anomaly scoring, while $\mathcal{B}_c$ stores
historical representations for context enrichment.

For an incoming chunk $\mathcal{C}_q$, context retrieval uses
only entries retained from earlier chunks. After prediction
and prototype adaptation on $\mathcal{C}_q$ are completed,
we update $\mathcal{B}_c$ using the original ego-graph
representations $\{\mathbf{z}_i:e_i\in E_q\}$.
We prefer representations that are representative of the
current chunk while remaining different from those already
stored in $\mathcal{B}_c$. Representativeness is measured
by the average similarity to other representations in the
same chunk. We define this measurement as the coverage score $\rho_i^{\mathrm{cov}}$:
\begin{equation}
\rho_i^{\mathrm{cov}}
=
\frac{1}{|E_q|-1}
\sum_{\substack{e_j\in E_q\\j\neq i}}
\mathrm{sim}(\mathbf{z}_i,\mathbf{z}_j),
\end{equation}
where $\mathrm{sim}(\cdot,\cdot)$ denotes a similarity
function, such as cosine similarity. The difference score
$\rho_i^{\mathrm{dif}}$ measures dissimilarity to the most similar representation
already stored in the buffer:
\begin{equation}
\rho_i^{\mathrm{dif}}
=
1-\max_{\mathbf{z}_j\in\mathcal{B}_c}
\mathrm{sim}(\mathbf{z}_i,\mathbf{z}_j).
\end{equation}
When $\mathcal{B}_c$ is empty, we set $\rho_i^{\mathrm{dif}}=1$,
so that initial candidate selection depends on coverage.

We assign each candidate a score computed by
$\rho_i^{\mathrm{cov}}\cdot\rho_i^{\mathrm{dif}}$ and select up to
10\% candidates from the current chunk.
Let $\mathcal{P}_q$ denote these candidates. For both new
candidates and existing buffer entries, we incorporate
recency into the retention score:
\begin{equation}
u_j^{(q)}=\rho_j^{\mathrm{cov}}\cdot\rho_j^{\mathrm{dif}}\frac{q_j}{q},
\end{equation}
where $q_j$ is the originating chunk ID of entry $j$.
The scores $\rho_j^{\mathrm{cov}}$ and $\rho_j^{\mathrm{dif}}$ are computed in chunk $q_j$,
using the representations in that chunk for coverage and the context buffer before that chunk is inserted for difference.
For each retained entry, we store these two scores and $q_j$ alongside its representation.
The stored scores remain fixed during subsequent retention decisions, while the recency factor $q_j/q$ is evaluated at the current chunk $q$.

We compare candidates and existing entries in the same pool
$\mathcal{U}_q=\mathcal{B}_c\cup\mathcal{P}_q$.
Their retention scores are min-max normalized over
$\mathcal{U}_q$ before ranking. The updated context buffer
retains the highest-ranked entries:
\begin{equation}
\mathcal{B}_c
\leftarrow
\operatorname{Top}_{\min(\mathcal{M}_{c,\max},
|\mathcal{U}_q|)}
\left(\mathcal{U}_q;u^{(q)}\right).
\end{equation}
Thus, each chunk contributes at most
10\% new entries, and the total buffer
size never exceeds $\mathcal{M}_{c,\max}$.
The updated buffer is used only for subsequent chunks.

For context retrieval, given a target edge $e_i$ in
$\mathcal{C}_q$ with representation $\mathbf{z}_i$,
we compute its similarity to the historical representations
in $\mathcal{B}_c$. We retrieve the
$\min(N_{con},|\mathcal{B}_c|)$ most similar entries
and denote their index set by $\mathcal{R}_i$.
The cross-chunk context vector is obtained by
recency-weighted aggregation:
\begin{equation}
\mathbf{c}_i
=
\frac{
\sum_{j\in\mathcal{R}_i}
\frac{q_j}{q}\mathbf{z}_j
}{
\sum_{j\in\mathcal{R}_i}\frac{q_j}{q}
}.
\end{equation}
Similarity determines which historical representations are
retrieved, while recency determines their aggregation weights.
The context vector is then fused with the current edge
representation:
\begin{equation}
\tilde{\mathbf{z}}_i
=
\lambda_c\mathbf{z}_i+
(1-\lambda_c)\mathbf{c}_i,
\end{equation}
where $\lambda_c$ controls the proportion of enrichment.
When the buffer is empty, including the first target chunk,
we directly set $\tilde{\mathbf{z}}_i=\mathbf{z}_i$.
The enhanced representations are used for anomaly scoring,
reliable pseudo-label selection, and prototype adaptation.
Current-chunk representations are added to $\mathcal{B}_c$
only after these steps, ensuring that context retrieval uses
historical information only.

Algorithm~\ref{alg:otta_dgad} summarizes source-domain
pretraining and online target-domain adaptation of OTTA-DGAD.
\begin{algorithm}[t]
\caption{OTTA-DGAD}
\label{alg:otta_dgad}
\small
\raggedright
\begin{algorithmic}[1]
\REQUIRE Labeled source graphs $\{\mathcal G_s\}_{s=1}^{S}$;
chronological unlabeled chunks $\{\mathcal C_q\}_{q=1}^{Q}$;
epochs $\mathcal I$; capacities $\mathcal M,\mathcal M_{c,\max}$;
integer budget $N_{con}$.
\STATE \textbf{Source pretraining (Sections~IV-A--IV-B)}
\STATE Initialize $\theta,\lambda_n,\lambda_a$ and
$\mu_n,\mu_a,\Sigma_n,\Sigma_a$; set $\mathcal B\leftarrow\emptyset$.
\FOR{each source graph $\mathcal G_s$}
    \FOR{each of $\mathcal I$ epochs}
        \STATE Initialize $(\mathbf v_n,\mathbf v_a)$ when
        $\mathcal B=\emptyset$; otherwise use its best retained pair.
        \STATE Encode source ego-graphs and optimize the
        scoring/alignment objective in Eqs.~\eqref{eq:encoder},
        \eqref{eq:scoring}--\eqref{eq:source-objective}.
        \STATE Rank by $s_d$ for the first source or $s_r$ for later
        sources; keep at most $\mathcal M$ highest-ranked pairs
        in $\mathcal B$ using Eqs.~\eqref{eq:separation}--\eqref{eq:retention}.
        \STATE Update $\mu_n,\mu_a,\Sigma_n,\Sigma_a$
        using Eq.~\eqref{eq:statistics}.
    \ENDFOR
\ENDFOR
\STATE Freeze $\theta,\lambda_n,\lambda_a$;
initialize $\mathcal B_c\leftarrow\emptyset$.
\FOR{each incoming target chunk $\mathcal C_q$}
    \STATE Encode each temporal ego-graph as $\mathbf z_i$
    using Eq.~\eqref{eq:encoder}.
    \STATE \textbf{Cross-Chunk Context Enrichment (Sec.~IV-D)}
    \IF{$\mathcal B_c=\emptyset$}
        \STATE Set $\tilde{\mathbf z}_i\leftarrow\mathbf z_i$.
    \ELSE
        \STATE Retrieve the $\min(N_{con},|\mathcal B_c|)$
        most similar entries and aggregate $\mathbf c_i$
        using Eq.~(21).
        \STATE Fuse $\mathbf z_i$ and $\mathbf c_i$
        using Eq.~(22).
    \ENDIF
    \STATE Score $\tilde{\mathbf z}_i$ using Eq.~\eqref{eq:scoring}
    and output $\{p_i:e_i\in E_q\}$ before any update.
    \STATE \textbf{Structure-Guided Prototype Updating (Sec.~IV-C)}
    \STATE Compute $\hat y_i$, entropy confidence, structural
    support, and $r_i$ using Eqs.~(9)--(13).
    \STATE Normalize scores within each present class using
    Eqs.~(14)--(15).
    \STATE Select the top $\min(N_{con},|E_q|)$ edges
    by $\tilde r_i$ and split them into $E_q^n,E_q^a$.
    \IF{$\sum_{e_i\in E_q^c}\tilde r_i>0$ for some $c\in\{n,a\}$}
        \STATE For each such class, aggregate its representative using
        Eq.~(16) and update its
        prototypes in $\mathcal B$ using Eq.~\eqref{eq:alignment};
        retain prototypes of all other classes.
        \STATE Update $\mu_n,\mu_a,\Sigma_n,\Sigma_a$
        using Eq.~\eqref{eq:statistics}.
    \ELSE
        \STATE Keep $\mathcal B$ and
        $\mu_n,\mu_a,\Sigma_n,\Sigma_a$ unchanged.
    \ENDIF
    \STATE Compute $\rho_i^{\mathrm{cov}}$ from $\mathbf z_i$ by
    Eq.~(17) when $|E_q|>1$; otherwise set $\rho_i^{\mathrm{cov}}=1$.
    Compute $\rho_i^{\mathrm{dif}}$ by Eq.~(18) when
    $\mathcal B_c\ne\emptyset$; otherwise set $\rho_i^{\mathrm{dif}}=1$.
    \STATE Select up to 10\% of current representations
    by $\rho_i^{\mathrm{cov}}\rho_i^{\mathrm{dif}}$ as candidates $\mathcal P_q$.
    \STATE Form $\mathcal U_q\leftarrow\mathcal B_c\cup\mathcal P_q$;
    compute $u_j^{(q)}$ using each entry's originating-chunk
    scores and chunk ID in Eq.~(19).
    \STATE Update $\mathcal B_c$ with the top
    $\min(\mathcal M_{c,\max},|\mathcal U_q|)$ entries from
    $\mathcal U_q$ using Eq.~(20); store each retained
    representation with its coverage/difference scores and originating chunk ID.
\ENDFOR
\end{algorithmic}
\end{algorithm}

\section{Experiments}

\subsection{Datasets}
We use 10 real-world datasets~\cite{liu2021anomaly, lee2024slade, yang2012defining} from various domains for evaluation, with detailed dataset statistics presented in Table~\ref{sta}. For datasets without ground-truth anomalies, we follow previous work~\cite{liu2021anomaly} and inject anomalies at random timestamps. Specifically, let $p$ denote the anomaly injection ratio and $m$ the total number of samples in the dataset. We link $p \times m$ previously disconnected nodes to create injected anomalies. 

\begin{table}[h]
    \caption{Statistics of datasets}
  \centering
  \scalebox{0.80}{
  \small
  \begin{tabular}{ c |c c c c}
\hline\hline
    \textbf{Dataset} & \textbf{Domain} & \textbf{\#Nodes} & \textbf{\#Edges} & \textbf{Avg. Degree} \\
\hline
Wikipedia & Online Collaboration & 9,227 &157,474&34.13 \\
Bitcoin-OTC & Transaction Network & 5,881 &35,588&12.10 \\
Bitcoin-Alpha & Transaction Network & 3,777  &24,173&12.80 \\
Email-DNC & Online Communication & 1,866&39,264&42.08\\
UCI Messages & Online Communication & 1,899&13,838&14.57\\
AS-Topology & Autonomous Systems &34,761&171,420&9.86\\
MOOC & Online Courses & 7,074&411,749 & 116.41\\
Synthetic-Hijack &Email Spam &986 &333,200 &38.11\\
TAX51 &Transaction Network&132,524&467,279&7.05\\
DBLP&Citation Network&25387& 119899&9.44\\
\hline\hline
    \end{tabular}}

\label{sta}
\end{table}

\subsection{Performance Metrics and Baselines}
We adopt AUROC and AUPRC as the evaluation metrics~\cite{liu2021anomaly,qiao2025anomalygfm,li2023zero}. Higher metric values indicate better detection performance.

Since AUROC and AUPRC are undefined when a chunk contains only one class, we exclude such chunks from chunk-level metric computation. Let $\mathcal{K}_{valid}$ denote the set of valid target chunks that contain at least one normal edge and one abnormal edge. The final online evaluation results are obtained by averaging the chunk-level metrics over all valid chunks:
\begin{equation}
\mathrm{AUROC}_{avg}
=
\frac{1}{|\mathcal{K}_{valid}|}
\sum_{k\in\mathcal{K}_{valid}}
\mathrm{AUROC}_k,
\end{equation}
\begin{equation}
\mathrm{AUPRC}_{avg}
=
\frac{1}{|\mathcal{K}_{valid}|}
\sum_{k\in\mathcal{K}_{valid}}
\mathrm{AUPRC}_k.
\end{equation}

This protocol evaluates the model in a strictly online manner, where adaptation and prediction are both performed sequentially along the target stream without revisiting full historical target chunks. We select the following baseline methods for comparison:
\begin{itemize}
    \item[$\bullet$] DP-DGAD \cite{zheng2026dp}:
    Learns normal and anomalous prototypes from labeled source
    graphs and updates its prototype memory online using
    high-confidence target pseudo-labels.

    \item[$\bullet$] GeneralDyG \cite{yang2025generalizable}:
    We add a transformer and a binary classification head for
    supervised temporal ego-graph learning, and update target-side
    prototypes during online adaptation.

    \item[$\bullet$] FALCON \cite{chen2024fine}:
    We add label-aware supervision with binary cross-entropy (BCE)
    loss and adapt its temporal representations using target
    pseudo-labels.

    \item[$\bullet$] SLADE \cite{lee2024slade}:
    We add a binary classification objective for source training
    and update its temporal self-supervised objective during
    target adaptation.

    \item[$\bullet$] SAD \cite{tian2023sad}:
    We combine its deviation loss with label-aware contrastive
    learning and update its memory bank and target representations
    online.

    \item[$\bullet$] TADDY \cite{liu2021anomaly}:
    We use a GNN--transformer encoder trained with BCE loss
    and adapted online using target pseudo-labels.

    \item[$\bullet$] ARC \cite{liu2024arc}:
    We add a transformer-based temporal encoder for dynamic
    graph modeling and adapt the resulting model online.

    \item[$\bullet$] AnomalyGFM \cite{qiao2025anomalygfm}:
    We add a temporal transformer and learn normal and anomalous
    prototypes from source data. During adaptation, only the
    normal prototype is updated; the source encoder and anomalous
    prototype remain fixed.

    \item[$\bullet$] UNPrompt \cite{niu2024zero}:
    We replace its backbone with a GNN--transformer encoder,
    pretrain it through dynamic graph contrastive learning,
    and update neighborhood prompts online.

    \item[$\bullet$] GraphPrompt \cite{liu2023graphprompt}:
    We use a GNN--transformer backbone, align source representations
    with normal and anomalous prototypes, and update prompts
    during target adaptation.

    \item[$\bullet$] ADCSD \cite{guo2025online}:
    We freeze the source-pretrained OTTA-DGAD backbone and
    replace context enrichment and prototype updates with
    lightweight short- and long-term score-correction modules
    adapted using target pseudo-labels.

    \item[$\bullet$] LCoTTA \cite{duan2026lifelong}:
    We freeze the source-pretrained DGAD backbone and update
    lightweight adaptation parameters by projecting pseudo-label
    gradients onto an online-tracked subspace of recent gradients,
    without context enrichment or prototype updates.
\end{itemize}
To evaluate models under continually evolving target distributions, we adopt an online test-time adaptation (OTTA) setting on chunked target dynamic graphs. Specifically, 1) the model is first pre-trained on multiple labeled source datasets and then adapted to an unlabeled target dataset. To comply with the OTTA setting, the source-pretrained backbone is frozen during target-time adaptation, and only lightweight adaptation modules are allowed to be updated. 2) Each target dynamic graph is chronologically divided into temporally ordered chunks, where each chunk contains a fixed number of target edges, set to 128 here. For each incoming chunk, the model first performs inference on the current chunk, and then updates its adaptive components using only the current chunk, without accessing any future information. 3) This test-then-adapt process is repeated sequentially over the whole target stream, and the final performance is reported by averaging over all target chunks.
\subsection{Experimental Settings}
All methods undergo pretraining on two source datasets, Wikipedia and MOOC~\cite{liu2021anomaly}, chosen because they have available ground-truth anomalies. Parameters for ego-graph extraction, GNN, and the transformer are aligned with those of GeneralDyG~\cite{yang2025generalizable}. The number of confident detections for pseudo-labeling, $N_{con}$, is 10\% of the chunk size. $\mathcal{M}_{c,\max}$ is set to 10\% of the size of the larger training source dataset, MOOC. The momentum $\alpha$ is set to 0.9. The value of $p$ is varied across 1\%, 5\%, and 10\%. The loss ratios $\lambda_A$ and $\lambda_{BCE}$ are set to 0.1 and 0.9, respectively, while $\lambda_d$ and $\lambda_{e}$ are set to 0.3 and 0.7. $\lambda_c$ is set to 0.5.

\begin{table*}[t]
\caption{Overall model performance under the OTTA setting across eight datasets with three anomaly percentages.}
\label{result_formatted_by_metric}
\renewcommand{\arraystretch}{1.1}
\centering
\setlength{\tabcolsep}{2pt}
\resizebox{\textwidth}{!}{%
\begin{tabular}{@{}ccl|ccc|ccc|ccc|ccc|ccc|ccc|ccc|ccc@{}}
\hline\hline
\multirow{2}{*}{\textbf{Metrics}} & \multirow{2}{*}{\textbf{Category}} & \multirow{2}{*}{\textbf{Methods}} & \multicolumn{3}{c|}{AS-Topology} & \multicolumn{3}{c|}{Bitcoin-Alpha} & \multicolumn{3}{c|}{Email-DNC} & \multicolumn{3}{c|}{UCI Messages} & \multicolumn{3}{c|}{Bitcoin-OTC} & \multicolumn{3}{c|}{TAX51} & \multicolumn{3}{c|}{DBLP} & \multicolumn{3}{c}{Synthetic-Hijack} \\
\cline{4-27}
& & & 1\% & 5\% & 10\% & 1\% & 5\% & 10\% & 1\% & 5\% & 10\% & 1\% & 5\% & 10\% & 1\% & 5\% & 10\% & 1\% & 5\% & 10\% & 1\% & 5\% & 10\% & 1\% & 5\% & 10\% \\
\hline

\multirow{13}{*}{\rotatebox{90}{AUROC}}
& \multirow{5}{*}{\rotatebox{90}{DGADs}}
&SAD & 40.16 & 43.81 & 43.43 & 57.61 & 55.59 & 60.27 & 50.05 & 54.54 & 52.28 & 56.91 & 58.28 & 59.74 & 70.04 & 69.19 & 74.12 & 42.52 & 43.56 & 46.39 & 43.02 & 42.98 & 47.87 & 59.85 & 59.67 & 58.14 \\
& &SLADE & 40.47 & 42.21 & 44.42 & 60.62 & 59.25 & 62.98 & 47.98 & 50.37 & 52.11 & 54.64 & 55.28 & 58.32 & 67.34 & 66.35 & 70.44 & 46.68 & 48.04 & 49.33 & 42.11 & 41.17 & 41.35 & 60.53 & 60.34 & 59.99 \\
& &FALCON & 46.73 & 50.24 & 53.09 & 57.13 & 61.48 & 66.76 & 76.81 & 72.71 & 75.02 & 59.36 & 57.05 & 62.22 & 67.37 & 68.43 & 70.16 & 45.27 & 48.79 & 50.28 & 44.41 & 46.30 & 47.11 & 59.06 & 60.17 & 65.55 \\
& &GeneralDyG & 41.99 & 43.51 & 45.80 & 62.26 & 63.03 & 65.09 & 50.93 & 52.63 & 54.78 & 50.33 & 53.50 & 58.91 & 70.91 & 71.77 & 73.84 & 52.79 & 50.58 & 50.42 & 48.33 & 47.34 & 46.78 & 56.78 & 53.47 & 54.02 \\
& &TADDY & 40.11 & 42.57 & 41.08 & 43.69 & 45.44 & 42.03 & 50.77 & 53.27 & 56.90 & 49.82 & 50.39 & 56.12 & 44.47 & 50.79 & 48.72 & 40.48 & 42.21 & 46.94 & 40.32 & 40.59 & 42.40 & 58.77 & 60.59 & 62.10 \\
\cline{2-27}

& \multirow{4}{*}{\rotatebox{90}{Generalists}}
&GraphPrompt & 48.23 & 50.66 & 52.32 & 46.08 & 46.24 & 43.48 & 67.59 & 70.11 & 72.31 & 59.93 & 58.29 & 58.74 & 48.36 & 50.07 & 52.25 & 51.24 & 53.33 & 56.22 & 58.62 & 57.15 & 61.35 & 68.58 & 74.42 & 70.96 \\
& & UNPrompt & 50.29 & 52.21 & 56.57 & 43.01 & 46.73 & 48.62 & 68.34 & 71.07 & 72.58 & 56.74 & 58.28 & 60.01 & 43.09 & 50.35 & 52.85 & 53.77 & 54.61 & 58.05 & 60.37 & 56.69 & 62.20 & 70.32 & 73.64 & 77.58 \\
& & AnomalyGFM & 52.13 & 53.90 & 57.84 & 46.23 & 47.01 & 49.56 & 70.62 & 72.39 & 74.16 & 60.03 & 62.44 & 59.59 & 52.80 & 51.22 & 50.45 & 54.97 & 55.71 & 56.87 & 60.24 & 57.83 & 59.27 & 72.52 & 78.07 & 76.35 \\
& & ARC & 53.68 & 54.13 & 60.10 & 40.59 & 42.62 & 43.51 & 70.52 & 75.19 & 72.33 & 58.05 & 60.40 & 60.21 & 47.75 & 52.82 & 48.19 & 47.73 & 51.01 & 58.22 & 60.97 & 57.59 & 62.32 & 73.06 & 72.14 & 75.61 \\
\cline{2-27}

& \multirow{4}{*}{\rotatebox{90}{OTTA}}
& ADCSD & 70.03 & 71.27 & 70.81 & 74.68 & 73.22 & 72.95 & 71.60 & 73.75 & 72.17 & 54.43 & 55.36 & 59.86 & 72.75 & 70.75 & 76.84 & 61.54 & 58.67 & 56.06 & 54.83 & 55.35 & 56.74 & 72.43 & 72.02 & 73.96 \\
& & LCoTTA & 69.47 & 71.13 & 72.21 & 73.33 & 71.65 & 71.32 & 69.66 & 73.84 & 72.21 & 54.27 & 55.15 & 59.85 & 70.38 & 69.41 & 68.48 & 61.20 & 60.25 & 57.23 & 55.92 & 55.50 & 57.32 & 72.36 & 71.97 & 70.66 \\

& & DP-DGAD (OTTA)& 69.48 & 71.04 & 71.99 & 62.51 & 65.34 & 69.77 & 70.21 & 74.42 & 71.09 & 59.86 & 55.22 & 59.69 & 76.74 & 71.30 & 68.59 & 55.68 & 60.33 & 57.62 & 60.41 & 54.30 & 57.66 & 70.81 & 70.62 & 69.36 \\

& & \textbf{OTTA-DGAD} & \textbf{73.48} & \textbf{74.04} & \textbf{76.99} & \textbf{77.88} & \textbf{75.57} & \textbf{75.47} & \textbf{79.88} & \textbf{84.12} & \textbf{81.77} & \textbf{63.51} & \textbf{64.99} & \textbf{69.75} & \textbf{80.94} & \textbf{79.85} & \textbf{78.61} & \textbf{65.90} & \textbf{65.72} & \textbf{66.06} & \textbf{62.05} & \textbf{59.92} & \textbf{63.25} & \textbf{81.33} & \textbf{81.10} & \textbf{83.72} \\

\hline

\multirow{13}{*}{\rotatebox{90}{AUPRC}}
& \multirow{5}{*}{\rotatebox{90}{DGADs}}
&  SAD & 5.26 & 11.15 & 11.87 & 5.12 & 7.45 & 10.16 & 10.88 & 13.16 & 25.96 & 10.47 & 14.75 & 24.85 & 6.91 & 11.42 & 17.30 & 4.76 & 7.03 & 13.58 & 3.62 & 6.84 & 9.02 & 9.72 & 14.89 & 16.66 \\
& &  SLADE & 5.34 & 10.26 & 11.52 & 5.49 & 6.05 & 10.97 & 10.95 & 18.06 & 23.23 & 10.47 & 16.72 & 23.19 & 5.57 & 9.11 & 12.02 & 4.75 & 6.02 & 11.71 & 2.61 & 5.79 & 10.01 & 10.68 & 15.05 & 19.50 \\
& &  FALCON & 7.01 & 11.37 & 11.91 & 5.99 & 7.13 & 12.87 & 14.39 & 20.06 & 32.07 & 12.85 & 20.39 & 27.79 & 6.06 & 10.77 & 13.14 & 9.23 & 8.19 & 14.20 & 4.44 & 7.56 & 8.47 & 12.38 & 16.13 & 20.52 \\
& &  GeneralDyG & 5.40 & 12.30 & 19.88 & 5.45 & 4.97 & 10.07 & 11.46 & 13.71 & 29.06 & 12.63 & 18.10 & 20.98 & 7.81 & 12.08 & 17.61 & 4.66 & 5.04 & 10.10 & 3.30 & 6.54 & 9.55 & 10.21 & 17.99 & 26.61 \\
& &  TADDY & 5.09 & 10.60 & 11.17 & 4.38 & 7.03 & 10.55 & 11.48 & 14.39 & 26.56 & 10.55 & 16.37 & 21.25 & 5.09 & 10.17 & 11.36 & 5.11 & 8.31 & 10.08 & 3.05 & 7.21 & 9.02 & 8.53 & 14.22 & 21.41 \\
\cline{2-27}

& \multirow{4}{*}{\rotatebox{90}{Generalists}}
& GraphPrompt & 10.39 & 11.61 & 19.10 & 7.46 & 10.07 & 11.11 & 18.18 & 24.32 & 35.26 & 17.06 & 21.27 & 26.72 & 5.10 & 10.94 & 15.52 & 9.97 & 11.02 & 12.18 & 6.58 & 8.18 & 10.03 & 10.25 & 14.33 & 25.85 \\
& & UNPrompt & 10.54 & 12.77 & 19.26 & 8.01 & 9.27 & 11.81 & 20.02 & 23.94 & 37.32 & 17.65 & 23.87 & 29.09 & 5.01 & 12.27 & 16.74 & 10.66 & 11.83 & 13.18 & 7.14 & 7.60 & 10.96 & 11.57 & 14.03 & 26.22 \\
& & AnomalyGFM & 11.74 & 13.23 & 20.03 & 7.64 & 12.98 & 11.33 & 21.15 & 27.43 & 39.54 & 21.01 & 23.99 & 32.89 & 5.14 & 11.63 & 17.97 & 11.12 & 13.73 & 12.28 & 7.36 & 9.67 & 11.09 & 11.42 & 12.97 & 30.92 \\
& & ARC & 12.05 & 18.66 & 24.34 & 10.16 & 10.77 & 12.56 & 14.53 & 27.50 & 38.29 & 17.94 & 23.31 & 30.12 & 7.03 & 12.58 & 16.48 & 9.60 & 10.23 & 13.64 & 7.27 & 8.36 & 10.21 & 13.72 & 15.55 & 26.38 \\
\cline{2-27}

& \multirow{4}{*}{\rotatebox{90}{OTTA}}
& ADCSD & 10.26 & 16.02 & 26.09 & 11.72 & 12.32 & 13.38 & 17.76 & 28.27 & 37.89 & 20.45 & 23.68 & 31.01 & 10.65 & 11.23 & 17.27 & 10.85 & 11.11 & 13.55 & 8.63 & 8.87 & 11.79 & 15.63 & 16.88 & 28.73  \\
& & LCoTTA & 10.36 & 13.27 & 23.49 & 11.56 & 10.07 & 12.98 & 14.43 & 27.17 & 38.43 & 16.49 & 27.77 & 30.98 & 10.62 & 11.17 & 17.13 & 11.85 & 11.13 & 13.74 & 7.68 & 9.89 & 11.75 & 14.47 & 13.67 & 29.40 \\

& & DP-DGAD (OTTA) & 10.46 & 18.38 & 23.05 & 12.49 & 13.17 & 13.78 & 19.99 & 29.63 & 36.25 & 25.83 & 27.11 & 33.77 & 9.52 & 10.06 & 14.24 & 10.79 & 12.33 & 18.64 & 7.65 & 10.86 & 10.14 & 13.27 & 15.20 & 27.08 \\

& & \textbf{OTTA-DGAD} & \textbf{12.38} & \textbf{23.32} & \textbf{31.68} & \textbf{17.67} & \textbf{16.08} & \textbf{15.30} & \textbf{23.95} & \textbf{36.47} & \textbf{40.28} & \textbf{30.47} & \textbf{30.74} & \textbf{40.84} & \textbf{11.84} & \textbf{12.97} & \textbf{18.18} & \textbf{12.80} & \textbf{15.07} & \textbf{23.56} & \textbf{9.67} & \textbf{12.88} & \textbf{16.07} & \textbf{17.52} & \textbf{20.47} & \textbf{32.37} \\
\hline\hline
\end{tabular}%
}
\end{table*}

\subsection{Experimental Results}

\begin{figure*}[htbp] 
	
	\centering
	
	\includegraphics[width=18cm]{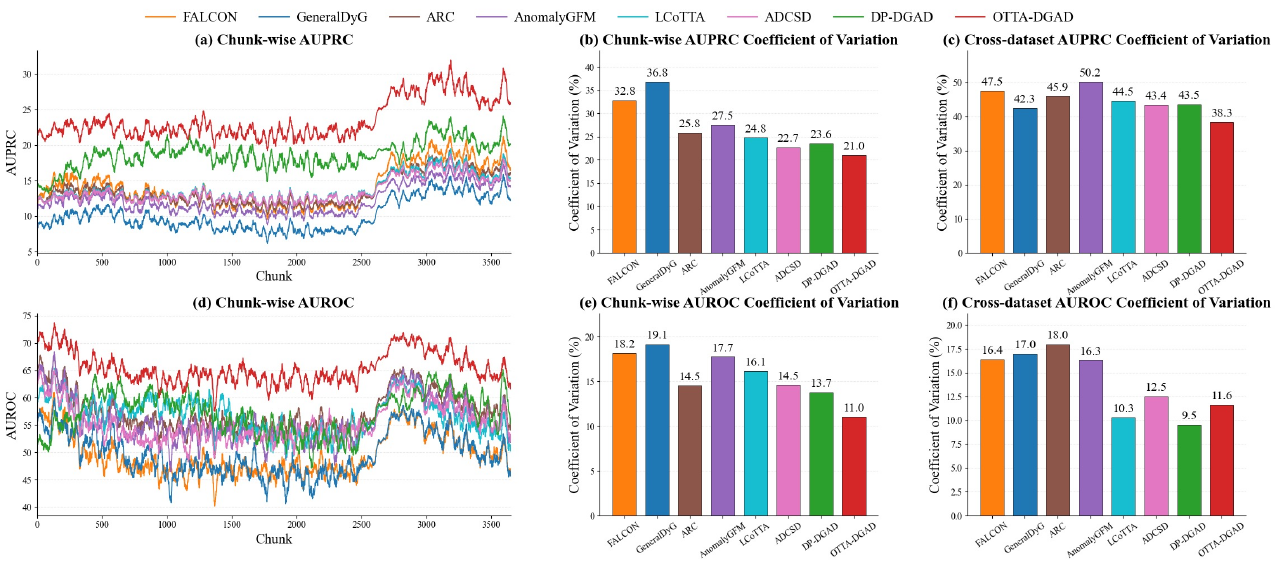}

	\caption{\textbf{Visualization of chunk-wise metrics and coefficients of variation across domains and chunks on TAX51 with a 10\% anomaly ratio.} }
	\label{stability}
	
\end{figure*}

In this section, we evaluate OTTA-DGAD's performance against three types of baselines: conventional DGAD methods without dedicated generalization modules, generalists with dedicated generalization modules, and OTTA models specifically designed for OTTA settings. We make the following observations: (1) OTTA-DGAD consistently performs better than all other baselines. DP-DGAD still performs well, but sometimes falls behind ADCSD and LCoTTA, whose dedicated OTTA modules make them better at online adaptation. Although generalists generally perform well, they cannot effectively incorporate domain-specific patterns, resulting in lower performance than OTTA models. Conventional DGAD methods perform the worst, mainly because of their limited generalization capabilities. (2) We use TAX51, which has the largest number of edges, with a 10\% anomaly ratio to visualize chunk-wise performance and analyze stability using the coefficient of variation across chunks and domains. As shown in Fig.~\ref{stability}(a) and (d), OTTA-DGAD consistently outperforms the other baselines. Fig.~\ref{stability}(b), (c), (e), and (f) also show that OTTA-DGAD is generally the most stable method across chunks and domains, while DP-DGAD and the other two OTTA baselines are among the next most stable methods.

In summary, these results demonstrate the state-of-the-art online performance and stability of OTTA-DGAD.

\begin{table*}[htbp]
\caption{Ablation results of different variants of OTTA-DGAD.}
\label{ablation}
\centering
\resizebox{\textwidth}{!}{%
\begin{tabular}{ll|cccccccc}
\hline\hline
\multirow{2}{*}{\textbf{Metric}} & \multirow{2}{*}{\textbf{Variants}} & \multicolumn{8}{c}{\textbf{Dataset}} \\
\cline{3-10}
& & AS-Topology & Bitcoin-Alpha & Email-DNC & UCI Messages & Bitcoin-OTC & TAX51 & DBLP & Synthetic-Hijack \\
\hline
\multirow{6}{*}{AUROC}
    & w/o DPAS           & 53.64& 62.21& 66.77& 58.90& 70.53& 44.11& 49.06& 60.12 \\

    & w/o ESGPU          & 63.34& 69.98& 72.50& 60.19& 74.57& 60.13& 55.58& 70.26  \\
    & w/o Wiki          & 60.01& 53.16& 67.21& 52.83& 48.04& 46.32& 50.28& 67.97\\
    & w/o MOOC          & 55.86& 60.07& 53.15& 60.37& 62.94& 56.36& 46.12& 58.81\\
    & w/o CCCE          & 63.38  & 62.53 & 67.04 & 61.13  & 68.76 & 54.25 & 60.84 & 68.71\\
    & w/o SGPU          & 66.47 & 67.81 & 70.26 & 63.65  & 69.11 & 60.08 & 60.30 & 73.49\\
 & \textbf{OTTA-DGAD} & \textbf{76.99} & \textbf{75.47}  & \textbf{81.77}  & \textbf{69.75}  & \textbf{78.61} & \textbf{66.06}  & \textbf{63.25} & \textbf{83.72}\\
\hline
\multirow{6}{*}{AUPRC}
    & w/o DPAS           & 19.73& 12.26& 32.15& 22.36& 16.14& 15.45& 9.01& 28.25 \\
    & w/o ESGPU          & 16.86& 10.05& 30.94& 30.92& 15.18& 17.54& 8.21& 27.07  \\
    & w/o Wiki          & 20.73& 10.25& 26.34& 20.58& 16.09& 10.21& 12.91& 25.18\\
    & w/o MOOC          & 17.22& 12.13& 20.69& 32.37& 12.19& 16.30& 10.06& 23.53\\
    & w/o CCCE          & 25.83  & 12.77 & 34.59 & 30.07  & 15.29  & 19.01  & 12.88 & 25.67 \\
    & w/o SGPU          & 27.15  & 11.56 & 31.09 & 30.14  & 12.87  & 20.55  & 10.11 & 26.24\\
   & \textbf{OTTA-DGAD}  & \textbf{31.68} & \textbf{15.30} & \textbf{40.28} & \textbf{40.84} & \textbf{18.18} & \textbf{23.56}  & \textbf{16.07} &  \textbf{32.37} \\
\hline\hline
\end{tabular}%
}
\end{table*}

\subsection{Ablation Study}
We evaluate the contributions of individual components through
six ablation variants:
(1) Dynamic Prototype-based Anomaly Scoring (DPAS) is replaced
with a standard binary classifier in \textbf{w/o DPAS}.
(2) Entropy-Structural Guided Prototype Update (ESGPU) is disabled
in \textbf{w/o ESGPU}, keeping the source-pretrained model and
prototype memory fixed throughout the target stream.
(3) \textbf{w/o Wiki} pretrains OTTA-DGAD on MOOC only.
(4) \textbf{w/o MOOC} pretrains OTTA-DGAD on Wiki only.
(5) Cross-Chunk Context Enrichment (CCCE) is removed in
\textbf{w/o CCCE}.
(6) Structure-Guided Prototype Updating (SGPU) is replaced
with entropy-only prototype adaptation in \textbf{w/o SGPU},
removing structural support from pseudo-label reliability
estimation.

Table~\ref{ablation} shows that OTTA-DGAD generally performs better than its ablated variants. Additionally, we have several observations: (1) The adaptation modules enable the model to better adapt to new domains while dynamic prototype updates allow it to adapt to individual chunks within each domain. Removing these modules leads to performance degradation. (2) Cross-chunk information is highly important, because it helps the model generalize across chunks and prevents it from forgetting important historical patterns. (3) Removing confident detection-guided adaptation prevents the model from capturing new patterns in an unlabeled target domain. However, entropy-based detection alone is insufficient, as w/o SGPU also suffers from performance degradation compared with the complete OTTA-DGAD model.

\begin{figure*}[htbp] 
	
	\centering
	
	\includegraphics[width=18cm]{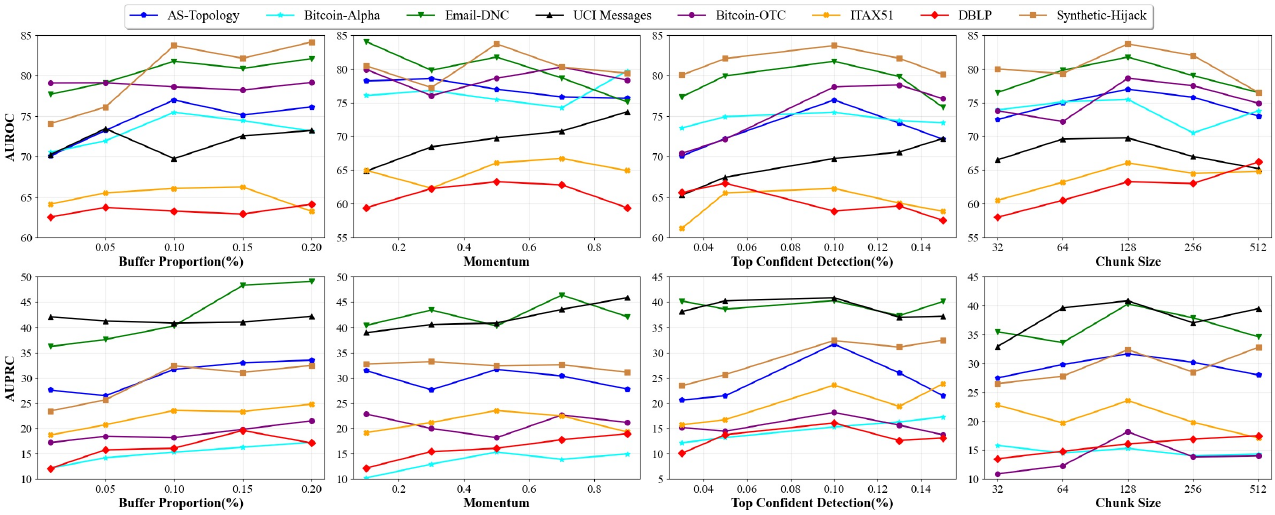}

\caption{\textbf{Impact of dynamic prototype memory buffer size $\mathcal{M}$, momentum $\alpha$, the number of confident detections $N_{con}$, and chunk size on OTTA-DGAD performance}. } \label{dpm}

\end{figure*}

\subsection{Hyperparameter Analysis}
To analyze OTTA-DGAD's sensitivity to hyperparameters, we examine four key parameters: the dynamic prototype memory buffer size $\mathcal{M}$, the momentum $\alpha$, the number of top confident detections $N_{con}$ (set to different proportions of the full dataset size here), and the chunk size processed at each step. We vary these values and observe their impact on model performance in Fig.~\ref{dpm}. Our findings are as follows. \textbf{(1) Increasing the memory buffer size generally improves performance.} 
This is because a larger memory buffer contains more prototype pairs, thus covering broader aspects of the distribution of abnormal patterns. However, we observe that larger buffer sizes actually degrade performance on some datasets. This occurs because oversized buffers may include irrelevant prototype pairs, which are neither similar to nor distinct enough from the new domains. They introduce bias and interfere with anomaly scoring, which relies on buffer prototypes to extract distributions. \textbf{(2) Using the top 10\% most confident detections appears to be the optimal setting.}
On most datasets, performance improves as the number of top confident detections increases. However, performance begins to drop when this percentage exceeds 10\%. This decline occurs because including more than the top 10\% of detections reduces reliability. These less confident detections have high entropy and introduce additional noise into the pseudo-labeling process. \textbf{(3) OTTA-DGAD remains stable under different momentum values.} We vary the momentum value and find that OTTA-DGAD remains stable across all datasets with no clear trend as momentum increases. This stability arises from the fact that the memory buffer already captures a sufficient number of similar abnormal and normal patterns under the constraint imposed by the selection score $s_r$. As a result, the extracted distributions remain stable regardless of $\alpha$'s value. \textbf{(4) A chunk size of 128 appears to be the optimal setting.} This result may arise because large chunk sizes, such as 256 and 512, can lead to overfitting to particular chunks, while small chunk sizes can lead to imbalance issues.

\begin{figure*}[htbp]
\centering
\includegraphics[width=\linewidth]{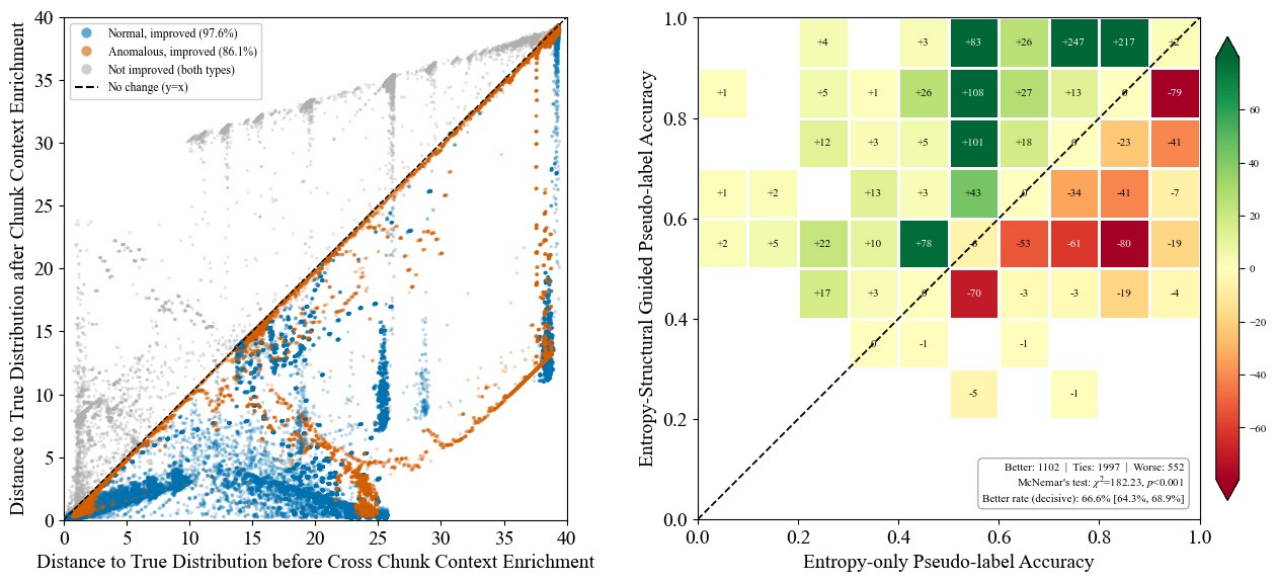}
\caption{\textbf{Effect of cross-chunk context enrichment and accuracy of structure-guided pseudo-labeling.} } \label{vis}
\end{figure*}

\subsection{Visualization}

In this section, we visualize representations enriched with cross-chunk context on TAX51 with a 10\% anomaly ratio and compare entropy-only pseudo-labeling with pseudo-labeling that additionally incorporates structural support.

It can be observed from the left panel of Fig.~\ref{vis} that 97.6\% of normal edges and 86.1\% of anomalous edges move closer to the true distribution after enrichment and more nodes fall to the bottom right of the diagonal line indicating no change.
Combining structural support with entropy-based confidence corrects cases where a prediction is confidently wrong but structurally inconsistent with its neighborhood. As shown in the right panel of Fig.~\ref{vis}, among the 1,654 chunks where the two criteria diverge, the model with structural support wins 66.6\% of the time, and McNemar's test confirms this advantage is highly significant, rather than attributable to chance. The large number of ties is expected since many chunks have limited anomaly samples.
\begin{figure*}[t]
\centering
\includegraphics[width=\textwidth]{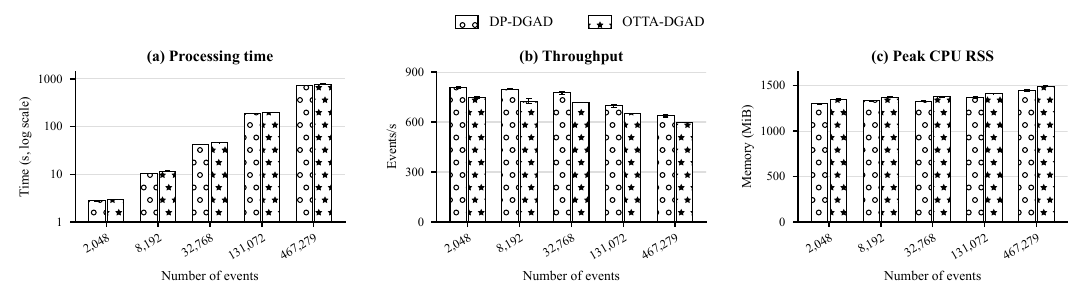}
\caption{\textbf{Scaling experiments on cumulative chunk-processing
time, throughput, and process-lifetime peak CPU RSS.}}
\label{fig:efficiency-scaling}
\end{figure*}
\subsection{Complexity Analysis}
\label{sec:complexity}

We compare OTTA-DGAD's storage with DP-DGAD and analyze the
computation cost introduced by our online modules.

\textbf{Storage.}
Let $B=|E_q|$ denote the chunk size and $d_p$ the representation
dimension. Under the same prototype memory capacity
$\mathcal{M}$, both methods retain at most $\mathcal{M}$
normal--anomalous prototype pairs in $\mathcal{B}$, together
with the class means $\mu_n,\mu_a$ and full covariance
matrices $\Sigma_n,\Sigma_a$. Let $S_{\mathrm{DP}}$ and
$S_{\mathrm{OTTA}}$ denote persistent storage for buffers
and distribution statistics in the two methods.
For DP-DGAD,
\begin{equation}
S_{\mathrm{DP}}=O(\mathcal{M}d_p+d_p^2).
\label{eq:dp-memory-complexity}
\end{equation}
OTTA-DGAD additionally stores at most
$\mathcal{M}_{c,\max}$ historical representations in the
context buffer $\mathcal{B}_c$, giving
\begin{equation}
S_{\mathrm{OTTA}}=
O\!\left(
(\mathcal{M}+\mathcal{M}_{c,\max})d_p+d_p^2
\right).
\label{eq:memory-complexity}
\end{equation}
Thus, the additional context storage is
$O(\mathcal{M}_{c,\max}d_p)$, including per-entry scores
and timestamps. Its capacity controls the retained
historical context without requiring storage to grow
with stream length. Actual occupancy may increase while
the buffer fills, up to this limit. These bounds exclude
model parameters, data-loader and graph-index storage,
and temporary computation tensors.

\textbf{Computation.}
The two methods share ego-graph encoding and
prototype-based anomaly scoring. Let
$b_q=|\mathcal{B}_c|\leq\mathcal{M}_{c,\max}$ be the
context buffer occupancy before chunk $q$. With direct
cosine similarities, CCCE requires $O(B^2d_p)$ for
within-chunk coverage and $O(Bb_qd_p)$ for historical
retrieval and difference scores. If full sorting is used,
retrieval ranking adds $O(Bb_q\log(1+b_q))$, and
retention-pool ranking adds
$O((b_q+B)\log(1+b_q+B))$.

For reliable pseudo-label selection, structural support
costs $O(\sum_{e_i\in E_q}|\mathcal{N}_i|)$, where
$\mathcal{N}_i$ is the neighboring-edge set defined in
Section~IV-C. Class-wise normalization costs $O(B)$,
while sorting-based reliable-edge selection costs
$O(B\log B)$, as in entropy-only ranking.
Reliability-weighted class aggregation costs at most
$O(Bd_p)$. 

In conclusion, OTTA-DGAD
uses additional storage and computation to provide
cross-chunk context and structure-guided adaptation.
Section~\ref{sec:efficiency} evaluates the resulting
trade-off between performance and measured
resource costs.


\subsection{Efficiency and Scalability}
\label{sec:efficiency}

We evaluate the efficiency of OTTA-DGAD against DP-DGAD on TAX51. We process chronological prefixes
containing the first 2,048, 8,192, 32,768, 131,072, and 467,279
edges, referred to here as events, with the last setting covering the complete stream.
Each prefix starts at the first event and is evaluated in an
independent process using three paired seeds, yielding 30 runs
in total. 
Cumulative chunk-processing time includes all chunks, whereas
throughput and latency exclude the first 1,024 warm-up events. 
P50 and P95 are the median and 95th-percentile chunk-processing
latencies within each run, characterizing typical and tail
processing costs, respectively. We report their means over
three paired runs. Higher throughput and lower latency indicate
greater processing efficiency.

\begin{table}[t]
\caption{Resource usage on the complete TAX51 stream. Values are means over three paired runs.}
\label{tab:efficiency-tax51}
\centering
\small
\setlength{\tabcolsep}{3pt}
\renewcommand{\arraystretch}{1.08}
\begin{tabular}{@{}lrr@{}}
\hline\hline
Metric & DP-DGAD & OTTA-DGAD\\
\hline
Throughput (events/s) & 636.24 & 597.45\\
P50 latency (ms/chunk) & 198.41 & 211.52\\
P95 latency (ms/chunk) & 241.28 & 254.54\\
Peak GPU allocated (MiB) & 208.05 & 208.05\\
Peak GPU reserved (MiB) & 470.00 & 470.00\\
Peak CPU RSS (MiB) & 1,442.76 & 1,495.71\\
\hline\hline
\end{tabular}
\end{table}

Table~\ref{tab:efficiency-tax51} characterizes the resource
trade-off introduced by cross-chunk context enrichment and
structure-guided adaptation. DP-DGAD achieves higher throughput
due to its simpler online processing, while OTTA-DGAD additionally
retrieves historical representations, evaluates structural
reliability, and maintains the context buffer. These additional
operations only incur a modest cost.

The latency results further indicate that the additional
processing cost remains limited for both typical and slower
chunks. P50 and P95 increase modestly, with no disproportionate
increase in tail latency. Therefore, the proposed
mechanisms improve online detection while introducing a
modest additional cost. 

Figure~\ref{fig:efficiency-scaling} indicates that OTTA-DGAD
maintains a scaling pattern similar to DP-DGAD while incorporating
historical context and structure-guided adaptation. Both methods
exhibit comparable throughput degradation as longer prefixes
are processed, and the additional mechanisms do not introduce
a progressively widening relative processing penalty over the
evaluated range. This suggests that their computational cost
remains manageable when extending detection to the complete
target stream. Reusing cached representations allows historical
information to support current detection without repeatedly
encoding previous chunks.

\section{Conclusion}
In this paper, we propose a generalist DGAD detector that achieves robust performance when target data arrive as online unlabeled chunks. Specifically, we aim to capture the evolving anomalous patterns that are both domain-agnostic and domain-specific, thereby achieving better generalizability than existing methods. To this end, we introduce OTTA-DGAD, a generalist detector that leverages dynamic prototypes extracted from temporal ego-graphs to model evolving anomalies. We further update the memory buffer using confident pseudo-labels, enabling effective adaptation to an unlabeled target domain. A cross-chunk context enrichment module is added to preserve historical evolving patterns. Extensive experiments across 10 real-world datasets from diverse domains validate the effectiveness of OTTA-DGAD.

\section*{Acknowledgment}
AI usage: The authors used Codex to assist with language editing of the manuscript. 

\normalem
\bibliographystyle{IEEEtran}
\bibliography{reference}

\end{document}